\RequirePackage{fix-cm}
\documentclass[twocolumn]{svjour3}          

\smartqed  
\usepackage{graphicx}
\usepackage{amsmath,amssymb,amsfonts}
\usepackage{algorithmic}
\usepackage{subcaption}
\usepackage{textcomp}
\usepackage{xcolor}
\usepackage{tabularx}
\usepackage[normalem]{ulem}
\usepackage[utf8]{inputenc}
\usepackage[numbers]{natbib}
\usepackage[hyphens]{url}

\begin{document}

\title{Practical and Configurable Network Traffic Classification Using Probabilistic Machine Learning
}

\author{Jiahui Chen \and Joe Breen \and Jeff M. Phillips \and Jacobus Van der Merwe       
}

\institute{J. Chen \at 
            \email{jiahui.chen@utah.edu}
            \and
          J. Breen \at \email{Joe.Breen@utah.edu}
            \and 
          J. Phillips \at \email{jeffp@cs.utah.edu}
            \and
          J. Van der Merwe \at \email{kobus@cs.utah.edu} 
}

\date{Received: date / Accepted: date}

\maketitle

\begin{abstract}
Network traffic classification that is widely applicable and highly accurate is valuable for many network security and management tasks.
A flexible and easily configurable classification framework is ideal, as it can be customized for use in a wide variety of networks.
In this paper, we propose a highly configurable and flexible machine learning traffic classification method that relies only on statistics of sequences of packets to distinguish known, or approved, traffic from unknown traffic.
Our method is based on likelihood estimation, provides a measure of certainty for classification decisions, and can classify traffic at adjustable certainty levels.
Our classification method can also be applied in different classification scenarios, each prioritizing a different classification goal.
We demonstrate how our classification scheme and all its configurations perform well on real-world  traffic from a high performance computing network environment.

\end{abstract}

\section{Introduction}
\label{intro}
Effective and practical classification of network traffic is crucial to many network management and security tasks. Categorization of network traffic yields valuable information on a network's activity, and timely classification enables this information to be quickly acted upon to ensure a secure and efficient network. Anomaly detection, quality of service monitoring, intrusion or attack detection, and resource allocation planning are all difficult network management tasks where traffic classification plays a critical role in solving \cite{sdn-attacks}. 
With the pervasive and diverse usage of the internet and online devices, large volumes of traffic from many different applications are constantly hosted on networks. Robust and flexible traffic classification is a difficult task due to the wide variety of traffic and dynamic nature of source applications. Traffic classification techniques have changed greatly over time, in reaction to changes in networking as a field.

Early and simple methods of traffic classification use port numbers to identify the traffic sources \cite{port-1, port-2, port-3}.
However as more applications used undisclosed, protocol-based, or configurable ports, port numbers became too unpredictable to be a reliable source of classification \cite{moore-towards, 2006-survey, clustering-erman}. 
In response to port-based classification becoming less effective, research turned to classification methods that use data packet inspection to find application or protocol signatures, i.e. patterns or data specific to the source application or protocol \cite{moore-towards, construction-of-app-sigs, p2p-app-sigs, blinc}. These methods require the ability to inspect packet payloads, so they are unable to classify encrypted traffic. Additionally, they are computationally expensive and require up-to-date application or protocol signatures to match traffic with \cite{intrustion-detection}. These issues present considerable limitations to inspection-based classification.

Most current approaches to traffic classification use machine learning algorithms and statistical properties of traffic flows to categorize traffic. A flow is usually defined by all packets with the same 5-tuple: source/destination IP, source/destination port, protocol. The statistical properties of flows, e.g., Inter-arrival time, Total Bytes, Average Packet Size, are referred to as features. Using statistical features of networking activity for classification avoids using port numbers or packet payloads, thereby remedying the limitations of the previously mentioned port and payload based methods. Machine learning techniques rely on the fact that different applications have differing networking behavior and patterns. These differences are represented in features, then discovered and used to discern flows' classes by a machine learning model.

In this paper, we focus on traffic in the Science DMZ \cite{science-dmz}, where we have a predominance of ``elephant flows" vs. ``mice flows"\cite{LAN200646}. We present a machine learning technique that uses the statistics of subflows, i.e. some subset of packets from a flow, to classify traffic with a measure of certainty. 
We classify traffic using probabilistic learning with likelihood estimation and adjustable certainty levels. This approach allows our method to classify traffic at higher or lower confidence levels, based on network preferences. 
This approach also allows network administrators to configure and use our classification so that it performs best on the most important traffic in their network. 

Our method can operate in three different classification scenarios: (1) classification performed with strict certainty thresholds resulting in known, unknown, and uncertain classification decisions; (2) classification with majority likelihood, eliminating any uncertain classification decisions; (3) incremental classification, where the classifier gathers information subflow by subflow, enabling the classifier to reach a classification decision as soon as possible. These different classification options along with the adjustable classification certainty level allows our technique to be easily customized to best fit a network's needs. 

We classify traffic into known and unknown classes. The known class consists of traffic from some group of applications approved for network usage, and the unknown class consists of traffic from any applications not in the known group. These class definitions fit well into real-world networks like the Science DMZ, and take advantage of the fact that networks with specific intended application usage usually allow applications with similar functions and behaviors. The broad definition of the unknown class allows it to include a huge array of diverse application traffic, so the variation between unknown traffic and known traffic is bound to be greater than the variation within the known traffic class. The known class will generally contain applications with similar functions and traffic, but the unknown class will include a huge variety of applications that have different behaviors from the known traffic. Our method successfully finds and utilizes these differences for classification via machine learning. 
This class scheme is also flexible since the known class can be defined with any set of applications, allowing network administrators to define a custom known class for their network with applications that are allowed for usage on their network. 
Thus, our technique is easily configured to fit a variety of network needs and is widely applicable to many real-world networks. 
This work makes these main contributions: 
\begin{trivlist}
    \item $\bullet$ We present a probabilistic machine learning method that classifies traffic with a measure of certainty. We describe how the certainty of classification decisions can be easily configured to yield different results. 
    \item $\bullet$ We show that our method can be applied in 3 different classification scenarios, each prioritizing a different classification goal. 
    \item $\bullet$ We demonstrate how our method and all of its configurations can be used to effectively classify traffic in the Science DMZ \cite{science-dmz} network setting. 
\end{trivlist}

\section{Background and Related Work}
Traffic classification techniques using machine learning comprise two main components: the representation of network traffic and the machine learning algorithm. Additionally, many  different classification schemes have been used. From the vast existing research, we present a brief overview of work relevant to ours.

\subsection{Existing Work on Network Traffic Representation}
Many different representations and statistical features of flows have been explored in previous work. Statistics on packet size, arrival times, and packet types have resulted in high classification accuracy when used with a wide variety of machine learning methods \cite{dl-survey, 2006-survey, internet-traffic-classification-demystified, blinc}. These features can be calculated over all the packets in an entire flow or on some series of packets sampled from the flow \cite{dl-survey, 2006-survey, subflow-1, subflow-2}. Research also exists on feature selection techniques which are used to reduce the number of features needed for classification and to find optimal features that result in the best classification performance \cite{internet-traffic-classification-demystified, feature-select-2012}. In these works, packet size statistics and discrete feature values were found to enable classification accuracy of 93\% and above for multiple machine learning algorithms \cite{internet-traffic-classification-demystified}. 

Calculating features over an entire flow is not ideal for timely classification, prompting more practical methods that classify sequences of packets in a flow. Using  features  on  only  the  first  few  packets  of  flows was found to yield reasonable classification results \cite{dl-survey, internet-traffic-classification-demystified}.
Earlier  work  also  found  that  using  a  sequence  of  packets, or  subflows, of  as  few  as  25  packets  can  result in classification precision and recall of above 95\% \cite{subflow-1}.
This subflow work was expanded upon by \cite{subflow-2}, finding that classification performance is not affected by the position of the subflow within the overall flow or the direction of the packets.
In \cite{subflow-1, subflow-2, rnn-cnn} the length of the subflow (value of $N$) results in a trade off between classification performance and processing requirements. They found that higher values of $N$ lead to better classification, but require more processing time and memory \cite{subflow-1, subflow-2, subglow-3}.

\subsection{Existing Work on Machine Learning Algorithms for Network Traffic Classification}
Many different machine learning algorithms have been used for traffic classification. Early work used traditional supervised learning methods that classify traffic into pre-defined classes, such as decision trees and Bayesian analysis techniques~\cite{2006-survey, bayesian, subflow-1, comparison}. These methods have been shown to perform classification at accuracy above 95\% on various sets of applications \cite{2006-survey, bayesian}. Unsupervised and semi-supervised learning methods, where traffic is grouped based on similarity rather than explicitly classified into a class, have also been explored in \cite{clustering-erman, semi-supervised, minetrac, robust, offline-semi, intrustion-detection, data-stream}. Clustering unlabelled or partially labelled traffic resulted in classification accuracy of 90-93\% \cite{clustering-erman, semi-supervised}. 

Recent methods have used deep learning, with supervised classification performed by convolutional neural networks and recurrent neural networks \cite{dl-survey, rnn-cnn}. Some other neural network methods have used unsupervised learning to learn traffic representations as well as how to imitate traffic, using auto-encoders and generative adversarial neural networks \cite{dl-survey}. Various architectures of neural nets used for classification have achieved high accuracy of up to 96\% \cite{rnn-cnn}.

\subsection{Network Traffic Classification Schemes}
Most of this existing work classifies traffic by mapping it to an application, application type, or protocol. A few classify traffic into known and unknown classes by discerning a specific, known application or group of applications from other traffic \cite{subflow-1, subflow-2, Baker}. Our work uses this latter scheme of known and unknown classification as it is less explored, more flexible, and widely applicable. In one setting, known traffic could be defined as a broad set of non-malicious activities for a well-protected, general usage network. But in another setting it might be a small set of specifically approved applications on a network designed for specialized uses only, like the Science DMZ. The flexibility of this known vs. unknown classification brings additional challenges, as our classification method must be robust enough to perform well on many different sets of known applications. 

In addition to addressing the more challenging task of classifying traffic into flexible known and unknown classes, we consider classification in the Science DMZ network setting which has not been previously explored. 
A Science DMZ is a subnetwork, usually part of a university network, that is configured and designed to optimize the usage of high-performance scientific computing applications \cite{science-dmz}. This network definition fits well with our known vs. unknown classification, as a Science DMZ is intended to host traffic from specific scientific computing applications and no other traffic. Our traffic dataset is sourced from the University of Utah's Science DMZ, which allows us to evaluate our method on realistic high-performance computing traffic. Our approach performs classification at or near 100\% accuracy on representative Science DMZ traffic. In addition, we evaluated our classification performance on a more challenging traffic dataset to show that our method generalizes well.

\section{Traffic Representation Methodology}
A series of network traffic statistics (e.g., Total Bytes, Standard Deviation of Packet Size, Largest Packet Size) forms a feature vector representation of network traffic; this feature vector representation is necessary in order to use machine learning algorithms to classify network traffic. In this section we discuss how we represent network traffic flows in our machine learning approach. 

\subsection{Use of Sub-flows}
Network traffic flows are composed of packets with the same 5-tuple: source/destination IP, source/destination port, and protocol. No existing work uses statistical properties of individual packets to classify flows, as single packets do not provide enough information for effective classification. A notable amount of existing work uses statistics on all packets in a flow to classify flows \cite{2006-survey, clustering-erman, semi-supervised, Baker}. However, using all packets in a flow for classification requires the flow to finish before it can be classified. Therefore, techniques that analyze all packets fail to stop flows of unapproved network activity from completing, making them less viable for real-world networks. Using all packets for classification also incurs high memory and computational costs, since flows can be long and data-intensive, especially in the large science dataset transfers seen in the Science DMZ. 

Because of the aforementioned issues with using single or all packets in a flow, our classification method uses subflows: some subset of $N$ packets taken from any point in a flow. The use of subflows was first introduced in \cite{subflow-1}. 
We use $N$-packet subflows to represent our traffic, where $N=\{25, 100, 1000\}$. These values of $N$ were discussed, experimented upon extensively, and found to be sufficient subflow lengths in \cite{subflow-1, subflow-2, subglow-3}, with the larger values of $N$ leading to better classification performance but requiring more processing time and memory. Our statistical features are calculated over each $N$-packet subflow and all of our flows are split into $N$-packet subflows for classification.

Using subflows gives our classification approach the additional advantage of being able to gather multiple data points per flow. Each subflow gives our classifier some statistical data on the overall flow, so it can use each subflow to increase or decrease certainty in a classification decision for the overall flow. Thus, our classification approach can gain valuable classification progress for each encountered subflow, and can make a decision on an overall flow when a certainty threshold is reached.

\subsection{Statistical Features of Traffic}
Selecting useful statistical features calculated over a series of packets to represent network traffic is crucial to effective machine learning.
Table 1 on page 71 of \cite{2006-survey} breaks down various network traffic statistics and groups them according to previously used machine learning approaches. We considered a broad set of statistics used in previous work that were found to achieve the best network traffic classification performance \cite{Baker, internet-traffic-classification-demystified, bayesian, blinc, clustering-erman}.

\begin{figure}
    \includegraphics[width=\linewidth]{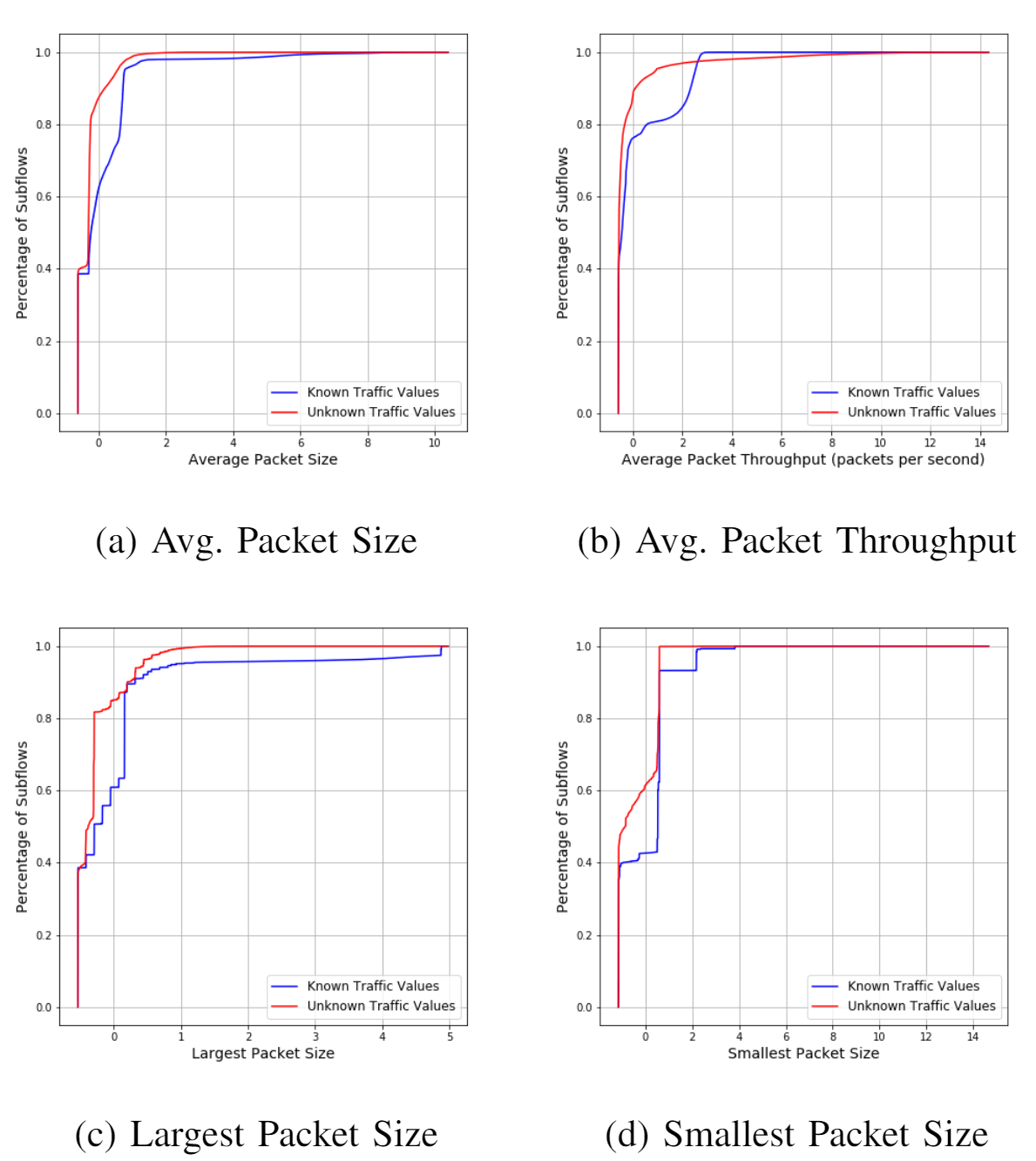}
    \caption{Feature Value CDFs for 100-Packet Subflows}
    \label{fig:feature-CDFs}
\end{figure}

To narrow down which features to use, we graphed the cumulative density function (CDF) of feature value distributions for our known and unknown traffic datasets to ensure that the features we use capture notable differences between known and unknown traffic. 
Fig. \ref{fig:feature-CDFs} shows example CDFs for various feature values. 

From our CDF analysis, we found that 14 of the following features effectively showed differences between known and unknown traffic: Total Bytes, Largest Packet Size, Smallest Packet Size, Number of TCP ACKs, Minimum Advertised Receive Window, Maximum Advertised Receive Window, Standard Deviation of Packet Size, Average Packet Size, Average Packet Inter-Arrival Time, Standard Deviation of Packet Inter-arrival Time, Maximum Packet Inter-arrival Time, Minimum Packet Inter-arrival Time, Average Packet Throughput (packets per second), Average Byte Throughput (bytes per second). 

Out of these 14 features, an even smaller subset of only 8 features were used in previous work that classified subflows to achieve high accuracies \cite{subflow-1, subflow-2}. Using a smaller number of features is favorable due to lower computational and memory costs; so we ran experiments using both sets of 14 and 8 features, to investigate whether or not using 14 features would yield performance gains that outweighed the higher computational cost. We found that using 14 features did not notably improve classification performance, so we used the following 8 features to represent our traffic: Maximum, Minimum, Mean, and Standard Deviation of Packet Inter-arrival Time and Packet Size. We calculate these 8 statistical features over all packets in each subflow; so each subflow is represented by an 8 element data point where each element is a feature value and is subsequently processed by our machine learning method as an 8-dimensional vector.

\section{Machine Learning Methodology}

\begin{figure}
    \centering
    \includegraphics[width=\linewidth]{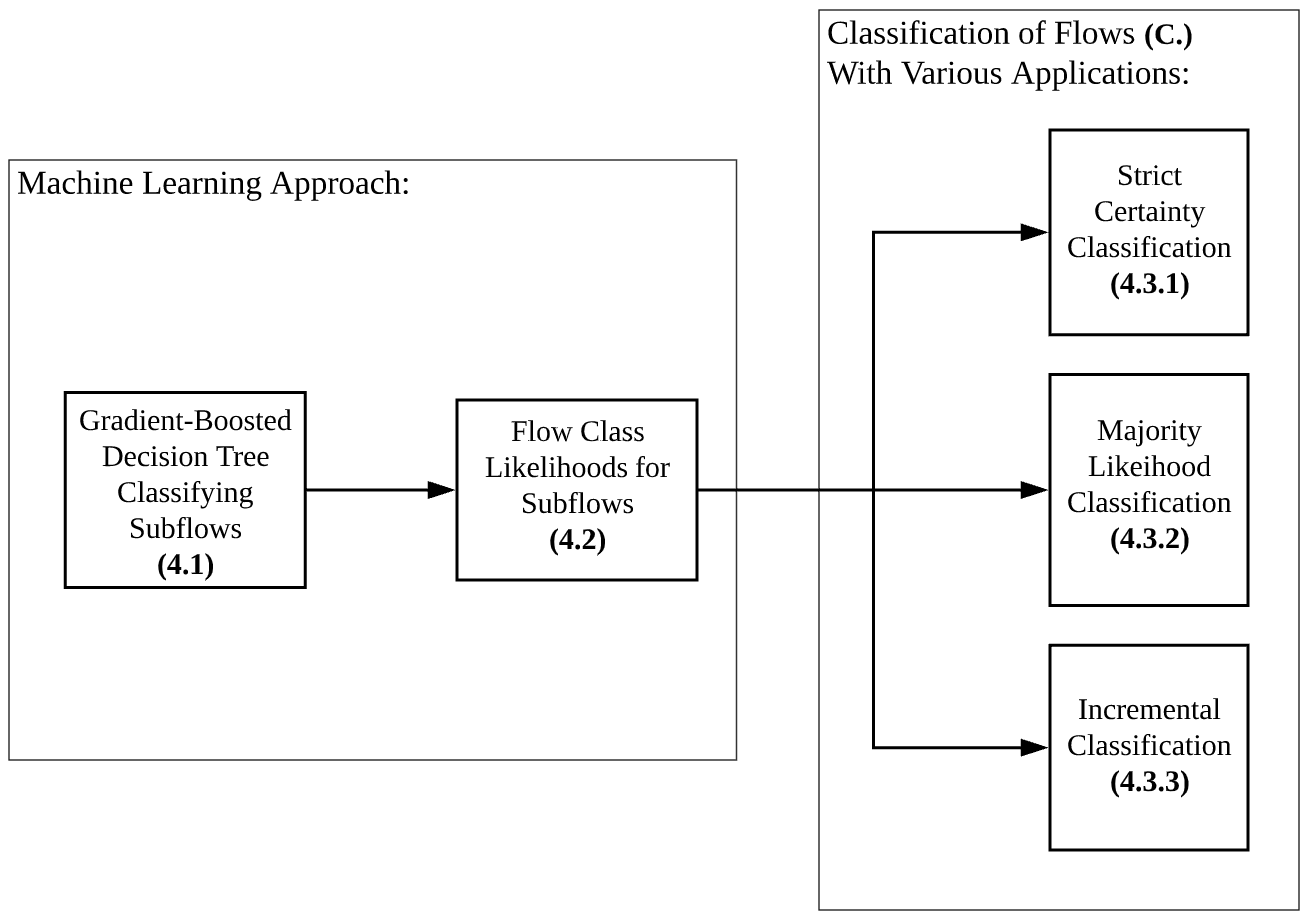}
    \caption{Machine Learning Approach and Applications (with corresponding paper sections)}
    \vspace{-0.1in}
    \label{method-flow}
\end{figure}

In this section we discuss the formulation and components of our machine learning approach as well as the different ways our classification method may be applied. 
Fig. \ref{method-flow} shows our methodology's components, pipeline, and multiple usage options.

\subsection{Classification of Individual Subflows}

Our machine learning approach classifies subflows, then utilizes the classification of individual subflows of a flow to classify the entire flow. We performed experiments comparing the subflow classification performance of Naive Bayes, Gradient Boosted Decision Tree (GBDT), Singular Vector Machine (SVM), and K-Nearest Neighbors (KNN) models, as these classifiers have been found to achieve high accuracies on traffic classification tasks in previous work \cite{bayesian, Baker, subflow-1, 2006-survey, comparison}. Gradient-boosted decision trees are known to be more powerful and robust than single decision trees \cite{GBDT}. For the SVM model, we use the one-class variant which has performed well on anomaly detection for networking traffic in previous work \cite{svm}. For our KNN experiments, we used $K=3$. 

\begin{table}
    \vspace{0.2in}
    \caption{Science DMZ Dataset Accuracies}
    \begin{tabularx}{\linewidth} { 
      | >{\centering\arraybackslash}X 
      | >{\centering\arraybackslash}X 
      | >{\centering\arraybackslash}X  
      | >{\centering\arraybackslash}X|}
     \hline
      \textbf{Classifier}: & \textbf{25-Packet-Subflows} & \textbf{100-Packet-Subflows} & \textbf{1000-Packet-Subflows} \\
     \hline
     Naive Bayes & 98.4 & 97.6 &  99.1 \\
     \hline
     Gradient-Boosted Decision Tree & 100 &  100 &  100 \\
    \hline
    One Class SVM & 28.2 & 36.7 & 81.4 \\
    \hline
    KNN & 100 & 100 & 100 \\
    \hline
    \end{tabularx}
    \label{science-dmz-indv}
    
     \vspace{0.1in}
    
    \caption{General Dataset Accuracies}
    \begin{tabularx}{\linewidth} { 
      | >{\centering\arraybackslash}X 
      | >{\centering\arraybackslash}X 
      | >{\centering\arraybackslash}X  
      | >{\centering\arraybackslash}X|}
     \hline
      \textbf{Classifier}: & \textbf{25-Packet-Subflows} & \textbf{100-Packet-Subflows} & \textbf{1000-Packet-Subflows} \\
     \hline
     Naive Bayes & 83.4 &  84.96 &  77.6 \\
     \hline
     Gradient-Boosted Decision Tree & 99.6 &  99.8 &  99.8 \\
    \hline
    One Class SVM & 67.9 & 68.6 & 68.5 \\
    \hline
    KNN & 98.3 & 98.7 & 98.4 \\
    \hline
    \end{tabularx}
    \label{general-indv}
\end{table}

Tables \ref{science-dmz-indv} and \ref{general-indv} show accuracies of all models for all subflow lengths, evaluated on test sets from both of our datasets; these test sets are held out from the data used to train these models.  All datasets and splits are described in more detail in Section 5.
Our results show that the GBDTs and KNN models perform very well on subflow classification, achieving accuracies above 98\%. Our ultimate goal is to classify entire flows rather than just individual subflows, so these high subflow classification accuracies serve as an important building block for our overall solution. 

Performing classification using KNN requires the calculation of distances between each data point to its $K$ nearest neighbors, which is much more computationally expensive than classification using the GBDT algorithm. Computational cost is especially important for practical network traffic classification approaches, as real-world flows can be large and real-time security actions based on classification decisions are ideal. Because the GBDT had the highest accuracies and is more computationally efficient than KNN, we use a GBDT model in the remainder of our machine learning framework.

\subsection{Establishing Flow Class Likelihoods From Individual Subflow Classification}

Our goal is to classify entire flows while only seeing subflows. The general idea is that each encountered subflow gives our classifier some statistical data on the overall flow; so the classifier can use each subflow to increase or decrease certainty in a classification decision for the overall flow.
We achieve this by assigning each subflow classification known and unknown flow class likelihoods. These flow class likelihoods can be thought of as estimated probabilities that a subflow belongs to an overall flow that is known or unknown, based on the subflow's label, or what the subflow is classified as. 
So we define and assign known and unknown flow class likelihoods to the known and unknown labels of subflows. 

In the set of training subflows $S$, each has a true label (known or unknown) and is given a predicted label (known or unknown).  On subflows outside of the training set, we only observe the predicted labels, so we can use the ratio of true labels to estimate the likelihood of the class on new data.  
Divide $S$ into $4$ sets: 
\begin{itemize}
    \item $S_k^k$ are from the known class and predicted as known,
    \item $S_u^k$ are from the known class and predicted as unknown,
    \item $S^u_k$ are from the unknown class and predicted as known, and
    \item $S^u_u$ are from the unknown class and predicted as unknown.  
\end{itemize}
We define flow class likelihoods in the following manner. Given a subflow is predicted as known, the sample likelihood it is actually known is:
$p_{k,k} = |S_k^k|/|S_k^k \cup S^k_u|$.
Similarly, the likelihood it is actually unknown is:
$p_{k,u} = |S_u^k|/|S_k^k \cup S^k_u|$.
For subflows predicted as unknown, we write the sample likelihood it is known as $p_{u,k} = |S_k^u|/|S_k^u \cup S^u_u|$ and the likelihood that it is unknown as $p_{u,u} = |S_u^u|/|S_k^u \cup S^u_u|$.  

With these class likelihoods associated with subflow labels, our machine learning approach can build up the likelihoods that a flow is known or unknown each time a subflow is encountered. In the next section, we explain in detail how these flow class likelihoods are utilized to classify flows.

\subsection{Classification Via Likelihood Estimation and Certainty Threshold}

We perform classification of a flow by combining the class likelihoods of a sequence of subflows belonging to that flow, using the class joint likelihoods of the subflows.
To create the class joint likelihoods over multiple subflows, we assume independence and take the product of all subflow likelihoods of the same class. These class joint likelihoods can be used as estimated probabilities that the sequence of subflows is of the corresponding class. The flow likelihoods can also be used to form a likelihood ratio, which we use as a measure of certainty for classification. The likelihood ratio is a fraction of the class likelihoods, indicating how much larger one class likelihood is than the other. For example, if the known class likelihood is 0.95 and the unknown class likelihood is 0.05 then the likelihood ratio is $\frac{0.95}{0.05}$. This indicates that under our model, we are 95\% certain that the flow is known, as the marginal probability that the flow is known, given all the subflows the classifier has seen, is $0.95$.  However, likelihoods of the numerator and denominator may not sum to $1$, and in general the joint ones will not.  But if the ratio is still $19$, e.g., $\frac{0.019}{0.001}$, then the confidence is still $95\%$.  

In particular, using our classifier, and these statistics, each subflow $s_j$ has a likelihood it is known $p_K(s_j)$ and a likelihood it is unknown $p_U(s_j)$.  These are defined based on the label:
\[
p_U(s_j) = \begin{cases} p_{u,u} & \text{ if } s_j \text{ labeled unknown } \\ 
                         p_{k,u} & \text{ if } s_j \text{ labeled known } \end{cases}
\]
and 
\[
p_K(s_j) = \begin{cases} p_{k,k} & \text{ if } s_j \text{ labeled known } \\ 
                         p_{u,k} & \text{ if } s_j \text{ labeled unknown. } \end{cases}
\]

We estimate the likelihood that a series of observed subflows $s_1, s_2, \ldots, s_m$ are known as:
\[
\hat L_K = p_{K}(s_1) \cdot p_{K}(s_2) \cdot \ldots \cdot p_{K}(s_m)
\]
and we use the same likelihood estimation for unknown $\hat L_U$ with the unknown likelihoods $p_U(s_j)$.

We define the likelihood ration as:
\[
\frac{\hat L_K}{\hat L_U} = \frac{p_{K}(s_1) \cdot p_{K}(s_2) \cdot \ldots \cdot p_K(s_m)}{p_{U}(s_1) \cdot p_{U}(s_2) \cdot \ldots \cdot p_U(s_m)}.
\]

By using a certainty threshold for classification, we can easily enforce the likelihood required for a flow to be classified. 
We enforce that $m \geq 15$, otherwise, because our subflow classifier has such high accuracy, it will always reach a $>95\%$ threshold after a single subflow.  

The use of different certainty thresholds for each class is also possible, which may be useful if the certainty of classification should be different between known and unknown traffic. For example, if a network is using our classification to block unknown traffic and wants to avoid disrupting allowed traffic, our technique would be applied with a very high certainty threshold for unknown classifications to ensure blocked traffic is classified as unknown with high confidence. The ease of adjusting classification certainty allows the certainty to be used as a parameter for classification. Different certainties can yield different classification accuracies depending on the underlying known and unknown traffic, and certainty can be a cross-validated hyperparameter that optimizes classification performance.

This likelihood estimation classification method can be applied in 3 different scenarios that we describe below and evaluate in our experiments:

\subsubsection{Strict Certainty Classification} 
In this classification scenario, flows are classified as known, unknown, or uncertain. If the known or unknown likelihood ratio reaches the desired certainty level, then the flow is classified as known or unknown. However, it is possible that neither likelihood ratio reaches the certainty level, so the flow is considered uncertain as its subflows do not yield a likelihood of high enough certainty for either class. Uncertain flows are indicative of traffic that is not similar enough to either class for a confident classification. 

This designation of uncertain flows may be useful as a means of filtering and monitoring traffic, enabling uncertain flows to be found and tracked. Uncertain flows may be used for further analysis with a more specific method of classification or inspected as the potential source of network issues. The amount of traffic classified as uncertain is configurable with the certainty level, as higher certainties result in more uncertain decisions.

\subsubsection{Majority Likelihood Classification}
For this classification scenario, if neither of the class likelihood ratios have reached the certainty level after all available subflows are seen, then the flow is classified as the class with the larger likelihood. This scenario results in no uncertain flow classifications since all uncertain flows are classified by their majority likelihood. This approach allows some flows to be classified with less certainty than the given certainty level, but generally increases accuracy in our experiments and is a viable option if uncertain flows are not desired.

\subsubsection{Incremental Classification}
In this classification scenario, the class likelihood ratios are updated with each encountered subflow's likelihoods, and classification occurs immediately once either class likelihood ratio reaches the given certainty level. Incremental classification takes full advantage of our usage of subflows, utilizing each sequence of packets in a flow to gain information on the flow and classify it after seeing the least amount of subflows possible. A classification decision is made as soon as possible, so this scenario prioritizes classification speed. In our Results section, we show that this scenario results in very fast classification after encountering a small fraction of subflows with excellent unknown detection capabilities. 
Note that incremental classification can use strict certainty or majority likelihood classification when making its classification decisions.

\section{Experiments And Results}

\subsection{Dataset}
To demonstrate and evaluate our classification method, we use the Science DMZ network. A Science DMZ is a security zone of a university campus network that is configured and designed to optimize the transfer of large scientific datasets \cite{science-dmz}. Researchers use the Science DMZ to transfer their datasets at high bandwith around the world, so a Science DMZ has performance-optimized security measures or other policy differences to enable faster data transfers. This networking environment fits well with our known vs. unknown classification, as a Science DMZ hosts traffic of specific scientific research applications and little other traffic. 

\begin{figure}[]
    \centering
    \includegraphics[width=\linewidth]{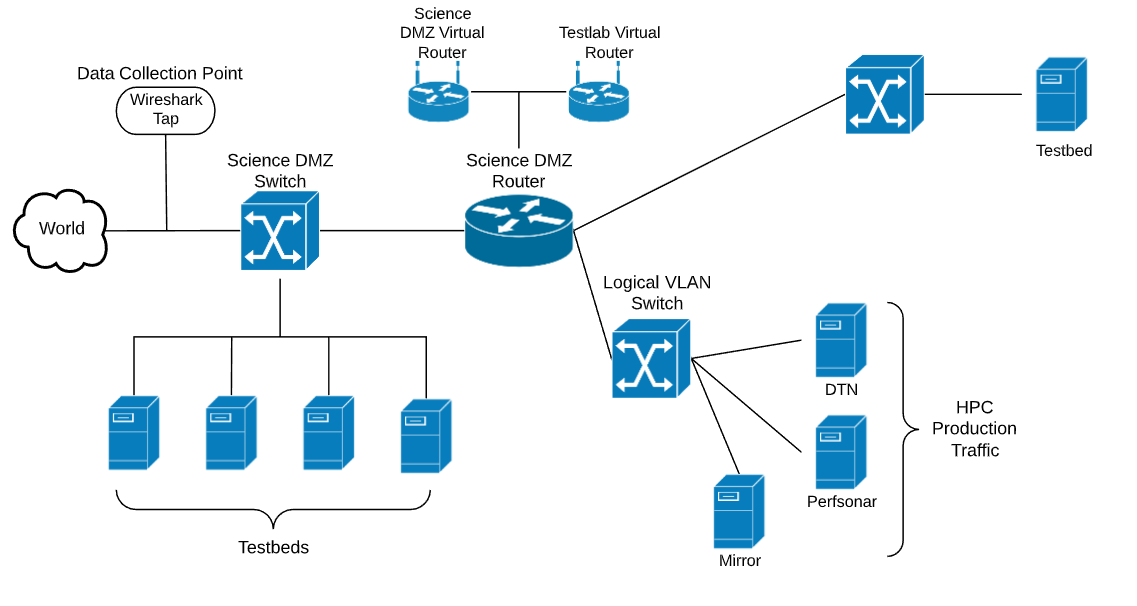}
    \caption{Data Collection Point in the University of Utah Science DMZ Sub-network}
    \vspace{-0.05in}
    \label{science-dmz}
\end{figure}

\begin{table}[]
    \caption{Dataset Statistics}
    \begin{tabularx}{\linewidth} { 
      | >{\centering\arraybackslash}X 
      | >{\centering\arraybackslash}X 
      | >{\centering\arraybackslash}X 
      | >{\centering\arraybackslash}X  
      | >{\centering\arraybackslash}X  
      | >{\centering\arraybackslash}X |}
     \hline
     & Globus & FDT & rclone & Mirror & WIDE \\
     \hline
     Bytes (GB)  & 51.6 & 129  & 82.1 & 42.6 & 30.48 \\
     \hline
     Flows & 185  & 72 & 12,292 & 2,239 & $1.112e6$ \\
    \hline
    \end{tabularx}
    \vspace{-0.15in}
    \label{data-table}
\end{table}

Fig. \ref{science-dmz} shows the location of our traffic capture tap in the University of Utah's Science DMZ, and Table \ref{data-table} shows size statistics of our dataset.
Note that we have different numbers of known and unknown flows, so our experimental accuracies are calculated separately for each label. All of our traffic is TCP and uses IPv4. We randomly select 80\% of our data for training and the rest for evaluation and ensured that the flows in the train and evaluation sets are mutually exclusive.

The specifics and application breakdowns of our known and unknown datasets are below.  

\subsubsection{Known Datasets}
Our known traffic is from 3 widely used large file transfer applications: Globus \cite{globus-1, globus-2}, FDT \cite{fdt}, and rclone \cite{rclone}.
We consulted domain experts and system administrators 
at the Center for High Performance Computing at the University of Utah 
to ensure that these 3 applications are commonly used by science researchers on the Science DMZ. The Globus captures were of ongoing file transfers between Globus endpoints at a university and various other universities in the United States. The FDT traffic was generated by moving DNA sequencing datasets from 
the Hunstman Cancer Institute to and from Data Transfer Nodes \cite{esnet-dtn} in the Science DMZ. The rclone traffic was generated by transferring ESnet test datasets \cite{esnet-test-data} to and from Google Drive. We verified with domain experts that our usage of FDT and rclone to generate traffic was consistent with their common usage in science research workflows, to ensure that our data is representative of real FDT and rclone traffic.

\subsubsection{Unknown Datasets}
For our unknown traffic, we use the Mirror and WIDE datasets. The Mirror dataset consists of random captures from a mirror server on the 
University of Utah's Science DMZ subnetwork that hosts repositories and other downloadable content. The WIDE dataset consists of captures, performed on the same dates as the Mirror captures, from the WIDE Traffic Archive \cite{WIDE}. The WIDE captures are from the main internet exchange link and internet service provider transit link of the WIDE organization \cite{WIDE}. 

\hfill

For all of our classification experiments, we train and evaluate our models using 2 different datasets. The known dataset always consists of the Globus, FDT, and rclone datasets but we use 2 different unknown class definitions: Science DMZ and General. The Science DMZ unknown class consists of only the Mirror traffic dataset, which was captured from 
the University of Utah's Science DMZ subnetwork but does not contain known application traffic. This approach allows us to simulate traffic classification in a realistic Science DMZ setting.
The General unknown class consists of both the Mirror and WIDE datasets, resulting in a much broader, more diverse unknown traffic class since WIDE's traffic is not from the same network and contains many more flows. Using this more varied unknown traffic allows us to evaluate how well our classification method generalizes when classifying more challenging, varied traffic.

\subsection{Strict Certainty Classification Results}

\begin{table}
   \caption{Science DMZ Dataset - Strict Certainty and Majority Likelihood Accuracies}
    \begin{tabularx}{\linewidth} { 
      | p{2cm}
      | >{\centering\arraybackslash}X 
      | >{\centering\arraybackslash}X 
      | >{\centering\arraybackslash}X  
      | >{\centering\arraybackslash}X|}
     \hline
 \textbf{Percentages of Subflows} & \textbf{25\%} & \textbf{50\%} & \textbf{75\%}  & \textbf{100\%}  \\
      \hline
     \multicolumn{5}{|c|}{Known Accuracies:} \\
     \hline
     25-Packet Subflows & 100 & 100 & 100 & 100 \\
     \hline
     100-Packet Subflows & 100 & 100 & 100 & 100 \\
    \hline
    1000-Packet Subflows & 100 & 100 & 100 & 100 \\
    \hline
    \multicolumn{5}{|c|}{Unknown Accuracies:} \\
    \hline
    25-Packet Subflows & 100 & 100 & 100 & 100 \\
     \hline
     100-Packet Subflows & 100 & 100 & 100 & 100 \\
    \hline
    1000-Packet Subflows & 100 & 100 & 100 & 100 \\
    \hline
    \end{tabularx}
    \label{table:mirror-strict}
\end{table}

To evaluate Strict Certainty classification, we perform our likelihood estimation classification and require a flow's class likelihood ratio to reach the given certainty threshold to be classified as known or unknown. Flows with class likelihood ratios that do not surpass the certainty threshold are considered uncertain. 
In our experiments, we perform classification using 25\%, 50\%, 75\%, and 100\% of subflows in each of the test set flows in order to evaluate classification performance when varying amounts of packets in flows are seen. Note that 100\% of subflows does not necessarily mean that all packets of the flow (from handshake to termination) are used, just that all captured packets of the flow are used. We use real-world datasets so, where it would be very limiting to only use completely captured flows for our experimental data. We require at least 15 subflows in a flow portion to perform classification. We also perform classification on features calculated over subflows of different packet lengths, using 25, 100, and 1000 packet subflows. 
We use these different combinations of percentage-defined subflow subsets and differing lengths of subflows to thoroughly evaluate classification in many situations where different portions of flows are seen.

\subsubsection{Science DMZ Dataset} 
Table \ref{table:mirror-strict} shows classification accuracies on the Science DMZ dataset, when using a strict certainty threshold of 95\%. 
Our accuracies are extremely high across all subflow sizes and subflow percentage subsets, with all experimental settings reaching 100\% accuracy. These results show that the unknown traffic is very different from the known application traffic and our method can successfully find and utilize these differences for classification. No flows were classified as uncertain across all experiments, even when requiring a high certainty threshold of 95\%. 

\subsubsection{General Dataset}
Fig. \ref{table:strict-general} shows classification accuracies on the General dataset, using the same strict certainty threshold of 95\%. 
Our accuracies are extremely high across all subflow sizes and subflow percentage subsets, with a minimum accuracy of 97.5\% and most experiments reaching 100\% accuracy.
These accuracies are slightly lower than the Science DMZ accuracies, which is expected since the General dataset contains unknown traffic that is more varied and similar to the known traffic, resulting in a more challenging classification task. 
For both known and unknown classificatoin, the 25-packet subflows had the poorest accuracies.
This indicates that our classification method performs better when subflows contain more packets, which makes sense as this means there's more networking traffic available for classification to be based on.
Only one experimental setting resulted in any flows that were unable to reach the 95\% certainty threshold necessary for classification, and thus considered uncertain. When performing classification on 50\% of 25-packets subflows, approximately 0.1\% of flows were considered uncertain.

\begin{table}
    \caption{General Dataset - Strict Certainty Accuracies}
\begin{tabularx}{\linewidth} { 
      | p{2cm}
      | >{\centering\arraybackslash}X 
      | >{\centering\arraybackslash}X 
      | >{\centering\arraybackslash}X  
      | >{\centering\arraybackslash}X|}
     \hline
 \textbf{Percentages of Subflows} & \textbf{25\%} & \textbf{50\%} & \textbf{75\%}  & \textbf{100\%}  \\
     \hline
     \multicolumn{5}{|c|}{Known Accuracies:} \\
     \hline
     25-Packet Subflows & 97.5 & 97.5 & 97.5 & 97.5 \\
     \hline
     100-Packet Subflows & 100 & 100 & 100 & 100 \\
    \hline
    1000-Packet Subflows & 100 & 100 & 100 & 100 \\
    \hline
    \multicolumn{5}{|c|}{Unknown Accuracies:} \\
    \hline
    25-Packet Subflows & 100 & 99.8 & 99.8 & 99.8 \\
     \hline
     100-Packet Subflows & 100 & 100 & 100 & 100 \\
    \hline
    1000-Packet Subflows & 100 & 100 & 100 & 100 \\
    \hline
    \end{tabularx}
    \label{table:strict-general}
\end{table}

Across both datasets, a very small amount of flows were considered uncertain even when a small percentage of subflows are seen. 
This shows that even if a high certainty for classification is enforced and not all packets in a flow are seen, our method can classify a majority of flows.

\subsection{Majority Likelihood Classification Results}
To evaluate Majority Likelihood classification, we perform our likelihood estimation classification to classify a flow as known or unknown if that flow's corresponding class likelihood ratio reaches the given certainty threshold. If after all available subflows are seen and the flow has no class likelihood ratio that has reached the certainty threshold, then the flow is classified as whichever class has the larger, or majority, likelihood estimate. 
We use the same percentage-defined subflow subsets and differing lengths of subflows as the Strict Certainty Classification experiments (25\%, 50\%, 75\%, and 100\% of a flow's subflows each with 25, 100, and 1000 packet subflows).

\subsubsection{Science DMZ Dataset}
For this dataset, all flows had class likelihood ratios that reached the 95\% certainty threshold across all percentages and sizes of subflows; so, no flows were considered uncertain and none needed to be classified using the majority class likelihood. This means that there are no differences in accuracy between Strict Certainty and Majority Likelihood classification for all experiments on the Science DMZ dataset, and Table \ref{table:mirror-strict} shows the unknown and known flow classification accuracies for Majority Likelihood classification.

\subsubsection{General Dataset}

\begin{table}
    \caption{General Dataset - Majority Certainty Accuracies}
    \begin{tabularx}{\linewidth} 
     { 
      | p{2cm}
      | >{\centering\arraybackslash}X 
      | >{\centering\arraybackslash}X 
      | >{\centering\arraybackslash}X  
      | >{\centering\arraybackslash}X|
    }
     \hline
 \textbf{Percentages of Subflows} & \textbf{25\%} & \textbf{50\%} & \textbf{75\%}  & \textbf{100\%}  \\
      \hline
     \multicolumn{5}{|c|}{Known Accuracies:} \\
     \hline
     25-Packet Subflows & 97.5 & 97.5 & 97.5 & 97.5 \\
     \hline
     100-Packet Subflows & 100 & 100 & 100 & 100 \\
    \hline
    1000-Packet Subflows & 100 & 100 & 100 & 100 \\
    \hline
    \multicolumn{5}{|c|}{Unknown Accuracies:} \\
    \hline
    25-Packet Subflows & 100 & 99.9 & 99.8 & 99.8 \\
     \hline
     100-Packet Subflows & 100 & 100 & 100 & 100 \\
    \hline
    1000-Packet Subflows & 100 & 100 & 100 & 100 \\
    \hline
    \end{tabularx}
    \label{table:general-maj}
\end{table}

Fig. \ref{table:general-maj} shows classification accuracies on the General dataset when using a certainty threshold of 95\%. Both the known and unknown accuracies do not notably differ from the Strict Certainty classification accuracies, as there were very few uncertain flows with class likelihood ratios that did not reach the 95\% threshold.
Using Strict Certainty classification, only the experimental setting using 50\% of 25-packet subflows resulted in uncertain flows; so only this experimental setting has a difference between the Strict Certainty and Majority Certainty classification accuracies.
It can be seen from the 0.1\% increase in accuracy from Strict Certainty classification that the 0.1\% of flows that were considered uncertain using Strict Certainty classification were classified correctly using Majority Certainty classification. This indicates that classification by the larger class likelihood is an effective way to classify traffic that is not similar enough to either class for a classification at the certainty required by the given threshold.

\subsection{Incremental Classification Results}
To evaluate Incremental classification, we update a flow's class likelihoods and check if the given certainty threshold is reached for every encountered subflow. Classification of the flow occurs immediately once a class likelihood reaches the certainty threshold, and subflows are encountered in chronological order of packet arrival; so flows are classified as soon as possible. 
Thus, these experiments allow us to evaluate how well our classification performs when reaching a classification decision in the fastest manner possible. We use both Strict Certainty and Majority Likelihood classification with this Incremental classification scheme, where Strict Certainty will allow for uncertain flows and Majority Likelihood will classify all flows as known or unknown even if no certain decision is reached after all subflows are seen. We evaluate on all lengths of 25, 100, and 1000 packet subflows. 

\subsubsection{Science DMZ Dataset}

\begin{figure}
    \centering
    \begin{subfigure}{.90\linewidth}
      \centering
      \includegraphics[width=\linewidth]{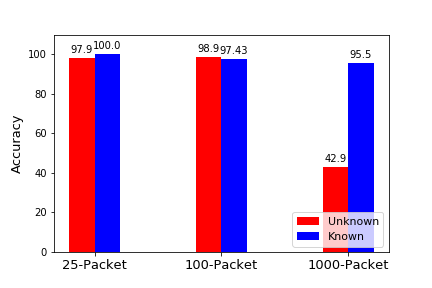}
      \caption{Strict Certainty Classification}
    \end{subfigure}
    \begin{subfigure}{.90\linewidth}
      \centering
      \includegraphics[width=\linewidth]{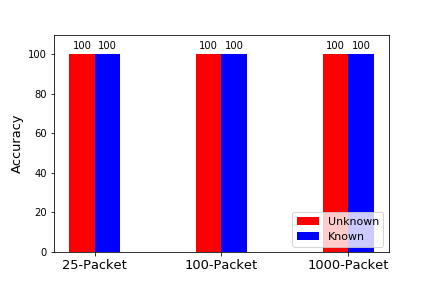}
      \caption{Majority Likelihood Classification}
    \end{subfigure}
    \caption{Science DMZ Dataset: Incremental Classification}
    \vspace{-0.1in}
    \label{fig:mirror-incremental}
\end{figure}

Fig. \ref{fig:mirror-incremental} shows classification accuracies on the Science DMZ dataset when using Incremental classification with both Strict Certainty and Majority Certainty classifications and a 95\% certainty threshold.
With Incremental classification, there are flows with class likelihood ratios that did not reach the certainty threshold, so the accuracies of Strict Certainty and Majority Likelihood classification notably differ.

Known accuracies of Strict Certainty classification are high across all subflow sizes, with a minimum accuracy of $95.5$\%. The average percentage of subflows needed to make a classification decision are overall very low: 1\% for 25-packet subflows, 5.4\% for 100-packet subflows, and 1.7\% for 1000-packet subflows. This shows that our method can classify known traffic very quickly with high accuracy, after seeing a very small percentage of packets or subflows. With Majority Likelihood classification, all known accuracies reach 100\%. This indicates that the small percentage of flows incorrectly classified by Strict Certainty classification were classified as uncertain and were correctly classified using Majority Likelihood classification.

Unknown accuracies of Strict Certainty classification for 25 and 100 packet subflows are around 98\%, but drop to 42.9\% for 1000-packet subflows. This accuracy drop is due to the 1000-packet subflow flows having considerably less subflows available for classification compared to the 25 and 100 packet subflow flows, since 1000-packet subflows require 10 times more packets per subflow than 100-packet subflows. This smaller number of subflows available for classification resulted in many flows being classified as uncertain, dropping the accuracy. With Majority Likelihood, all unknown accuracies reach 100. This indicates that classifying uncertain flows that did not reach the required certainty threshold by their majority class likelihood is an effective approach. 
These results show that classifying traffic using majority likelihood is a viable and simple option that enables improved classification accuracy and the elimination of uncertain flows.

\subsubsection{General Dataset}

\begin{figure}
    \centering
    \begin{subfigure}{.90\linewidth}
      \centering
      \includegraphics[width=\linewidth]{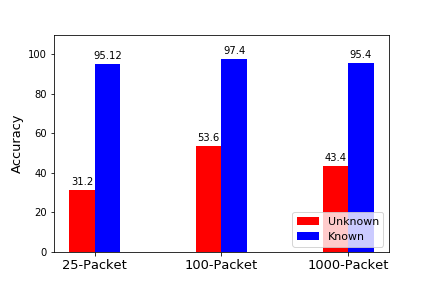}
      \caption{Strict Certainty Classification}
    \end{subfigure}
    \begin{subfigure}{.90\linewidth}
      \centering
      \includegraphics[width=\linewidth]{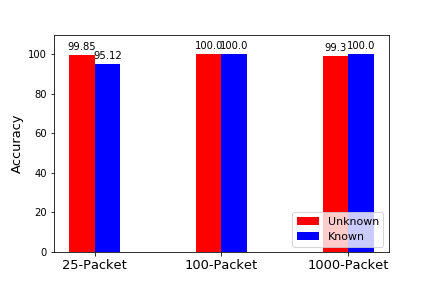}
      \caption{Majority Likelihood Classification}
    \end{subfigure}
    \caption{General Dataset: Incremental Classification}
    \vspace{-0.1in}
    \label{fig:general-incremental}
\end{figure}

Fig. \ref{fig:general-incremental} shows classification accuracies on the General dataset from Incremental classification with both Strict Certainty and Majority Likelihood classifications and a 95\% certainty threshold.

Known accuracies of Strict Certainty classification are high across all subflow sizes, with a minimum accuracy of $95.12$\%. The average percentage of subflows needed to make a classification decision for all subflow sizes were in the 1-5\% range, very low and similar to the percentages on the Science DMZ dataset. This shows that even on a more difficult dataset, our method is very effective at classifying known traffic as quickly as possible. Majority Likelihood classification either improves known accuracies to 100\% or does not change accuracy. 

Unknown accuracies from Strict Certainty classification are low but this is remedied by using Majority Likelihood classification, where accuracies reach a minimum of 99.85\%. This indicates that the low Strict Certainty classification accuracies are due to a considerable portion of flows being incorrectly classified as uncertain, but with Majority Likelihood classification these flows can be correctly classified. 
These results further support the viability of Majority Likelihood classification as a method to improve classification performance, especially when there are many uncertain flows from Strict Certainty classification.
The average percentage of subflows needed to make a classification decision is 41\% for 25-packet subflows, 43.2\% for 100-packet subflows, and 53\% for 1000-packet subflows; so overall a classification decision was made before half of available subflows were seen. 

Across both datasets, Incremental classification has better performance on known traffic than unknown traffic.
A classification can be made quickly after seeing less than half of subflows for both traffic classes, reinforcing our conclusions from Strict Certainty and Majority Likelihood that our method can classify a flow correctly after seeing a small portion of the flows' packets. Classifications of known traffic were made especially quickly, after seeing only 1-5\% of subflows, at accuracies above 95\%. This indicates our method can correctly classify known flows very quickly with minimal computation, as it only needs to process a tiny percentage of packets before making a classification. For unknown traffic, incremental classification using a strict certainty threshold of 95\% yields many uncertain flows. When Majority Likelihood classification is used, unknown flows that were considered uncertain with Strict Certainty classification can be correctly classified.

\section{Conclusion and Future Work}

In this paper, we introduced a machine learning method that uses statistics on sequences of packets, called subflows, to classify networking traffic as known or unknown with a measure of certainty. 
Our technique uses a gradient-boosted decision tree-based subflow classifier to assign class likelihoods to subflows, then uses joint likelihood estimations over multiple subflows to classify entire flows at a customizable certainty threshold.  

This method of classification allows traffic to be classified at an easily configurable certainty threshold and in three different ways. If used with Strict Certainty thresholds, flows are only classified as known or unknown if they can be classified at the given certainty level, and our method can find uncertain flows that are not similar enough to either class. If used with Majority Likelihood, all flows are classified as known or unknown by allowing some flows to be classified with whichever class likelihood estimate is higher rather than strictly requiring the certainty level. If used in an Incremental classification manner, each subflow updates the flow's class likelihood estimate and classification of a flow occurs after seeing the fewest number of subflows possible. 

We evaluated our technique on traffic from the Science DMZ subnetwork domain \cite{science-dmz}, as it naturally fits our class scheme and has not been used as a traffic classification setting before. We also evaluate on a more general, challenging dataset to ensure that our method can generalize well.
Our results show that our classification performs very well in the Science DMZ setting, able to reach 100\% accuracy for all classification options. On the general dataset, we maintained high accuracy on known traffic classification, reaching up to 100\%, though unknown classification accuracies dropped in the Strict Certainty classification scenario. 

Our method was shown to perform well even when only seeing a small percentage of flows, reaching accuracies up to 100 on both datasets when only a fourth of available subflows in a flow are used for classification. With Strict Certainty classification, very few flows are considered uncertain even when requiring 95\% certainty and seeing partial flows. The use of Majority Likelihood classification was shown to correctly classify flows deemed uncertain in Strict Certainty classification, improving classification performance. The Incremental classification approach reached classification decisions very quickly after seeing small amounts of subflows and maintained high accuracies on known flows across both datasets.

Our experiments show that in a real-world Science DMZ, our method is effective at classifying known and unknown traffic very quickly. With Incremental classification, accuracies above 95\% were reached after encountering as little as 1\% of subflows. Our Strict Certainty and Majority Likelihood results indicate that for all subflow sizes there's no drop in performance between the different percentages of subflows used for classification. This indicates our method can classify a flow well without needing to see a certain percentage of the flow's packets. Strict Certainty is able to correctly classify flows that reach the given certainty threshold and identify uncertain flows. 
If uncertain flows are not desired in a network setting, then Majority Likelihood classification can be used effectively; as it had extremely high accuracy across all experiments, even when used with Incremental classification. Incremental classification with Majority Likelihood has high performance and can classify known flows after seeing a tiny amount of subflows, meaning classification requires minimal time and computation.

Out of all the classification scenarios, Incremental classification accuracies dropped the most between the Science DMZ and General dataset results, so further work could be done to achieve more generalizable Incremental classification performance. 
In Incremental classification unknown accuracies are also generally lower than known accuracies, especially on the more difficult dataset. 
Maintaining high performance on unknown traffic classification is an expected challenge, as we define unknown traffic as any traffic that is not from the known applications, so unknown traffic can have huge variety. The most challenging datasets and network settings would have unknown traffic that has similar function and behavior as known traffic. Future work applying our approach to more challenging datasets could explore more sophisticated subflow classifiers, different formulations of class likelihoods of subflow classifications, or the use of regularization on class likelihoods to maintain high accuracies.

\hfill
\hfill
\hfill

\section*{Declarations} 

\subsection*{Funding}
This work was funded by NSF Awards ACI-1642158 and IIS-1816149.  

\subsection*{Conflicts of Interest}
None

\subsection*{Availability of Data and Material}
The datasets used are available here: \url{https://hive.utah.edu/concern/datasets/nk322d36m?locale=en}

\subsection*{Code availability}
The experimentation code is available here: \url{https://anonymous.4open.science/r/d6b2b12b-dfd4-4d45-98d9-d404caf02797/}

\subsection*{Informed Consent Statement}
Not applicable

\subsection*{Author Contribution}
Jiahui Chen is first author and implemented all experimentation as well as completed the majority of the writing. All the other authors supervised this work equally.

\bibliographystyle{spbasic}      
\bibliography{bib.bib}

\begin{thebibliography}{37}
\providecommand{\natexlab}[1]{#1}
\providecommand{\url}[1]{{#1}}
\providecommand{\urlprefix}{URL }
\expandafter\ifx\csname urlstyle\endcsname\relax
  \providecommand{\doi}[1]{DOI~\discretionary{}{}{}#1}\else
  \providecommand{\doi}{DOI~\discretionary{}{}{}\begingroup
  \urlstyle{rm}\Url}\fi
\providecommand{\eprint}[2][]{\url{#2}}

\bibitem[{LAN(2006)}]{LAN200646}
 (2006) A measurement study of correlations of internet flow characteristics.
  Computer Networks 50(1):46--62,
  \doi{https://doi.org/10.1016/j.comnet.2005.02.008}

\bibitem[{Allen et~al(2012)Allen, Bresnahan, Childers, Foster, Kandaswamy,
  Kettimuthu, Kordas, Link, Martin, Pickett, and Tuecke}]{globus-2}
Allen B, Bresnahan J, Childers L, Foster I, Kandaswamy G, Kettimuthu R, Kordas
  J, Link M, Martin S, Pickett K, Tuecke S (2012) Software as a service for
  data scientists. Commun ACM 55(2):81–88, \doi{10.1145/2076450.2076468},
  \urlprefix\url{https://doi.org/10.1145/2076450.2076468}

\bibitem[{Baker et~al(2018)Baker, Quinn, Phillips, and Van~der Merwe}]{Baker}
Baker R, Quinn R, Phillips J, Van~der Merwe J (2018) Toward classifying unknown
  application traffic. In: Proceedings. DYnamic and Novel Advances in Machine
  Learning and Intelligent Cyber Security DYNAMICS’18

\bibitem[{{Casas} et~al(2011){Casas}, {Mazel}, and {Owezarski}}]{minetrac}
{Casas} P, {Mazel} J, {Owezarski} P (2011) Minetrac: Mining flows for
  unsupervised analysis semi-supervised classification. In: 2011 23rd
  International Teletraffic Congress (ITC), pp 87--94

\bibitem[{Cho et~al(2000)Cho, Mitsuya, and Kato}]{WIDE}
Cho K, Mitsuya K, Kato A (2000) Traffic data repository at the wide project.
  In: USENIX 2000 FREENIX Track, USENIX

\bibitem[{Craig-Wood(2020)}]{rclone}
Craig-Wood N (2020) {rclone}. \urlprefix\url{https://rclone.org/}

\bibitem[{Erman et~al(2006)Erman, Arlitt, and Mahanti}]{clustering-erman}
Erman J, Arlitt M, Mahanti A (2006) Traffic classification using clustering
  algorithms. In: Proceedings of the 2006 SIGCOMM Workshop on Mining Network
  Data, ACM, New York, NY, USA, MineNet '06, pp 281--286,
  \doi{10.1145/1162678.1162679},
  \urlprefix\url{http://doi.acm.org/10.1145/1162678.1162679}

\bibitem[{Erman et~al(2007{\natexlab{a}})Erman, Mahanti, Arlitt, Cohen, and
  Williamson}]{offline-semi}
Erman J, Mahanti A, Arlitt M, Cohen I, Williamson C (2007{\natexlab{a}})
  Offline/realtime traffic classification using semi-supervised learning.
  Performance Evaluation 64:1194--1213, \doi{10.1016/j.peva.2007.06.014}

\bibitem[{Erman et~al(2007{\natexlab{b}})Erman, Mahanti, Arlitt, Cohen, and
  Williamson}]{semi-supervised}
Erman J, Mahanti A, Arlitt M, Cohen I, Williamson C (2007{\natexlab{b}})
  Semi-supervised network traffic classification. SIGMETRICS Perform Eval Rev
  35(1):369–370, \doi{10.1145/1269899.1254934},
  \urlprefix\url{https://doi.org/10.1145/1269899.1254934}

\bibitem[{ESnet(2020{\natexlab{a}})}]{esnet-test-data}
ESnet (2020{\natexlab{a}}) {ESnet Data Transfer Nodes}.
  \urlprefix\url{https://fasterdata.es.net/performance-testing/DTNs/}

\bibitem[{ESnet(2020{\natexlab{b}})}]{science-dmz}
ESnet (2020{\natexlab{b}}) {Science-DMZ}.
  \urlprefix\url{http://fasterdata.es.net/science-dmz/}

\bibitem[{ESnet(2020{\natexlab{c}})}]{esnet-dtn}
ESnet (2020{\natexlab{c}}) {Science DMZ: Data Transfer Nodes}.
  \urlprefix\url{https://https://fasterdata.es.net/science-dmz/DTN/}

\bibitem[{{Foster}(2011)}]{globus-1}
{Foster} I (2011) Globus online: Accelerating and democratizing science through
  cloud-based services. IEEE Internet Computing 15(3):70--73

\bibitem[{Friedman(2000)}]{GBDT}
Friedman JH (2000) Greedy function approximation: A gradient boosting machine.
  Annals of Statistics 29:1189--1232

\bibitem[{Haffner et~al(2005)Haffner, Sen, Spatscheck, and
  Wang}]{construction-of-app-sigs}
Haffner P, Sen S, Spatscheck O, Wang D (2005) Acas: Automated construction of
  application signatures. In: Proceedings of the 2005 ACM SIGCOMM Workshop on
  Mining Network Data, Association for Computing Machinery, New York, NY, USA,
  MineNet ’05, p 197–202, \doi{10.1145/1080173.1080183},
  \urlprefix\url{https://doi.org/10.1145/1080173.1080183}

\bibitem[{Karagiannis et~al(2005)Karagiannis, Papagiannaki, and
  Faloutsos}]{blinc}
Karagiannis T, Papagiannaki K, Faloutsos M (2005) Blinc: Multilevel traffic
  classification in the dark. In: Proceedings of the 2005 Conference on
  Applications, Technologies, Architectures, and Protocols for Computer
  Communications, Association for Computing Machinery, New York, NY, USA,
  SIGCOMM ’05, p 229–240, \doi{10.1145/1080091.1080119},
  \urlprefix\url{https://doi.org/10.1145/1080091.1080119}

\bibitem[{Karmakar et~al(2019)Karmakar, Varadharajan, and
  Tupakula}]{sdn-attacks}
Karmakar K, Varadharajan V, Tupakula U (2019) Mitigating attacks in software
  defined networks. Cluster Computing 22, \doi{10.1007/s10586-018-02900-2}

\bibitem[{Kim et~al(2008)Kim, Claffy, Fomenkov, Barman, Faloutsos, and
  Lee}]{internet-traffic-classification-demystified}
Kim H, Claffy K, Fomenkov M, Barman D, Faloutsos M, Lee K (2008) Internet
  traffic classification demystified: Myths, caveats, and the best practices.
  In: Proceedings of the 2008 ACM CoNEXT Conference, ACM, New York, NY, USA,
  CoNEXT '08, pp 11:1--11:12, \doi{10.1145/1544012.1544023},
  \urlprefix\url{http://doi.acm.org/10.1145/1544012.1544023}

\bibitem[{{Lopez-Martin} et~al(2017){Lopez-Martin}, {Carro},
  {Sanchez-Esguevillas}, and {Lloret}}]{rnn-cnn}
{Lopez-Martin} M, {Carro} B, {Sanchez-Esguevillas} A, {Lloret} J (2017) Network
  traffic classifier with convolutional and recurrent neural networks for
  internet of things. IEEE Access 5:18,042--18,050

\bibitem[{Moore and Papagiannaki(2005)}]{moore-towards}
Moore AW, Papagiannaki K (2005) Toward the accurate identification of network
  applications. In: Dovrolis C (ed) Passive and Active Network Measurement,
  Springer Berlin Heidelberg, Berlin, Heidelberg, pp 41--54

\bibitem[{Moore and Zuev(2005)}]{bayesian}
Moore AW, Zuev D (2005) Internet traffic classification using bayesian analysis
  techniques. In: Proceedings of the 2005 ACM SIGMETRICS International
  Conference on Measurement and Modeling of Computer Systems, Association for
  Computing Machinery, New York, NY, USA, SIGMETRICS ’05, p 50–60,
  \doi{10.1145/1064212.1064220},
  \urlprefix\url{https://doi.org/10.1145/1064212.1064220}

\bibitem[{Moore et~al(2001)Moore, Keys, Koga, Lagache, and Claffy}]{port-3}
Moore D, Keys K, Koga R, Lagache E, Claffy KC (2001) The coralreef software
  suite as a tool for system and network administrators. In: Proceedings of the
  15th USENIX Conference on System Administration, USENIX Association, USA,
  LISA ’01, p 133–144

\bibitem[{Newman(????)}]{fdt}
Newman H (????)

\bibitem[{t.~{Nguyen} and {Armitage}(2006)}]{subflow-1}
t~{Nguyen} TT, {Armitage} G (2006) Training on multiple sub-flows to optimise
  the use of machine learning classifiers in real-world ip networks. In:
  Proceedings. 2006 31st IEEE Conference on Local Computer Networks, pp
  369--376, \doi{10.1109/LCN.2006.322122}

\bibitem[{{Nguyen} and {Armitage}(2008{\natexlab{a}})}]{subglow-3}
{Nguyen} TTT, {Armitage} G (2008{\natexlab{a}}) Clustering to assist supervised
  machine learning for real-time ip traffic classification. In: 2008 IEEE
  International Conference on Communications, pp 5857--5862

\bibitem[{{Nguyen} and {Armitage}(2008{\natexlab{b}})}]{2006-survey}
{Nguyen} TTT, {Armitage} G (2008{\natexlab{b}}) A survey of techniques for
  internet traffic classification using machine learning. IEEE Communications
  Surveys Tutorials 10(4):56--76, \doi{10.1109/SURV.2008.080406}

\bibitem[{Nguyen et~al(2012)Nguyen, Armitage, Branch, and Zander}]{subflow-2}
Nguyen TTT, Armitage GJ, Branch P, Zander S (2012) Timely and continuous
  machine-learning-based classification for interactive ip traffic. IEEE/ACM
  Transactions on Networking 20:1880--1894

\bibitem[{Rezaei and Liu(2018)}]{dl-survey}
Rezaei S, Liu X (2018) Deep learning for encrypted traffic classification: An
  overview. CoRR abs/1810.07906,
  \urlprefix\url{http://arxiv.org/abs/1810.07906}, \eprint{1810.07906}

\bibitem[{Saroiu et~al(2003)Saroiu, Gummadi, Dunn, Gribble, and Levy}]{port-1}
Saroiu S, Gummadi KP, Dunn RJ, Gribble SD, Levy HM (2003) An analysis of
  internet content delivery systems. SIGOPS Oper Syst Rev 36(SI):315–327,
  \doi{10.1145/844128.844158},
  \urlprefix\url{https://doi.org/10.1145/844128.844158}

\bibitem[{{Sen} and {Jia Wang}(2004)}]{port-2}
{Sen} S, {Jia Wang} (2004) Analyzing peer-to-peer traffic across large
  networks. IEEE/ACM Transactions on Networking 12(2):219--232

\bibitem[{Sen et~al(2004)Sen, Spatscheck, and Wang}]{p2p-app-sigs}
Sen S, Spatscheck O, Wang D (2004) Accurate, scalable in-network identification
  of p2p traffic using application signatures. In: Proceedings of the 13th
  International Conference on World Wide Web, Association for Computing
  Machinery, New York, NY, USA, WWW ’04, p 512–521,
  \doi{10.1145/988672.988742},
  \urlprefix\url{https://doi.org/10.1145/988672.988742}

\bibitem[{{Shafiq} et~al(2016){Shafiq}, {Yu}, {Laghari}, {Yao}, {Karn}, and
  {Abdessamia}}]{comparison}
{Shafiq} M, {Yu} X, {Laghari} AA, {Yao} L, {Karn} NK, {Abdessamia} F (2016)
  Network traffic classification techniques and comparative analysis using
  machine learning algorithms. In: 2016 2nd IEEE International Conference on
  Computer and Communications (ICCC), pp 2451--2455

\bibitem[{Sun et~al(2019)Sun, Zhang, Yin, Wang, and Min}]{data-stream}
Sun R, Zhang S, Yin C, Wang J, Min S (2019) Strategies for data stream mining
  method applied in anomaly detection. Cluster Computing 22(2):399–408,
  \doi{10.1007/s10586-018-2835-2},
  \urlprefix\url{https://doi.org/10.1007/s10586-018-2835-2}

\bibitem[{Wang et~al(2020)Wang, Liang, Sun, and Ma}]{intrustion-detection}
Wang Y, Liang Y, Sun H, Ma Y (2020) Intrusion detection and performance
  simulation based on improved sequential pattern mining algorithm. Cluster
  Computing 23, \doi{10.1007/s10586-020-03129-8}

\bibitem[{Winter et~al(2011)Winter, Hermann, and Zeilinger}]{svm}
Winter P, Hermann E, Zeilinger M (2011) Inductive intrusion detection in
  flow-based network data using one-class support vector machines. In: NTMS,
  IEEE, pp 1--5,
  \urlprefix\url{http://dblp.uni-trier.de/db/conf/ntms/ntms2011.html#WinterHZ11}

\bibitem[{Zhang et~al(2012)Zhang, Lu, Qassrawi, Zhang, and
  Yu}]{feature-select-2012}
Zhang H, Lu G, Qassrawi MT, Zhang Y, Yu X (2012) Feature selection for
  optimizing traffic classification. Comput Commun 35(12):1457–1471,
  \doi{10.1016/j.comcom.2012.04.012},
  \urlprefix\url{https://doi.org/10.1016/j.comcom.2012.04.012}

\bibitem[{{Zhang} et~al(2015){Zhang}, {Chen}, {Xiang}, {Zhou}, and
  {Wu}}]{robust}
{Zhang} J, {Chen} X, {Xiang} Y, {Zhou} W, {Wu} J (2015) Robust network traffic
  classification. IEEE/ACM Transactions on Networking 23(4):1257--1270

\end{thebibliography}

\end{document}